\documentclass{article}

\usepackage{arxiv}

\usepackage[utf8]{inputenc} 
\usepackage[T1]{fontenc}    
\usepackage{hyperref}       
\hypersetup{
	colorlinks=true,
	linkcolor=blue,  
	urlcolor=blue,   
	citecolor=blue   
}
\usepackage{url}            
\usepackage{booktabs}       
\usepackage{amsfonts}       
\usepackage{nicefrac}       
\usepackage{microtype}      
\usepackage{lipsum}		
\usepackage{graphicx}
\usepackage{natbib}
\usepackage{doi}

\usepackage{bm} 
\usepackage{newtxtext}       %
\usepackage[varvw]{newtxmath}       

\title{Price Stability in the European Union: \\ A Systemic Approach Using Random Matrix Theory}

\author{ 
		{Sami Diaf} \\
	Department of Socioeconomics\\
	Universität Hamburg\\
	\texttt{sami.diaf@uni-hamburg.de} \\
}

\date{}

\renewcommand{\headeright}{Technical Report}
\renewcommand{\undertitle}{Technical Report}

\hypersetup{
pdftitle={Price Stability: A Systemic Approach Using Random Matrix Theory},
pdfsubject={q-bio.NC, q-bio.QM},
pdfauthor={Sami Diaf},
pdfkeywords={Inflation, Eurozone, Random Matrix Theory},
}

\begin{document}
\maketitle

\begin{abstract}
Price stability remains a pillar in monetary policy practices and carries a special importance within monetary unions. Mainstream economics tried to leverage price stability using price indices and several metrics to shed light on specific dynamics and optimal macroeconomic levels. The wide availability of data led researchers to consider the study of systems using Random Matrix Theory, based on inner correlation patterns. This aims to enhance the multivariate analysis by removing noisy patterns from the signal and improve data quality for further inferences. This work considers the collection of monthly inflation indices in the Eurozone as a \textit{system} of prices to analyze its eigenvalues' statistical and asymptotic properties and uncover inner country-level insights. Results confirm the system cannot assumed to be randomly generated, and the data exhibit noise-dominated patterns, due to small and persistent variations at the country-level. The latter make the inter-country correlations more dynamic and the separation of the signal from the noise quiet difficult. Findings identified two countries as distorting inflation dynamics besides three other distinct, regional-based groups of countries. Variability sources might stem from economic episodes fueling inflation spikes in some countries, as well as methodological aspects used to ensure data quality and representativeness in the European Union. Despite being complex, the system demonstrates a certain stability, in terms of self-organization; while large monthly fluctuations cannot be considered as rare events, but part of the data-generating process. 
\end{abstract}

\keywords{Inflation \and Eurozone \and Random Matrix Theory}

\section{Introduction}\label{section1}

Price stability has been debated and extensively researched by economists, aiming at finding the optimal level of prices to implement monetary policies. In this context, stability refers to an optimal price level that depends on the policy horizons \citep{Smets_2003_2}, which was widely thought to be the ultimate objective of monetary policy across several countries \citep{Castelnuovo_2003}.

In the case of monetary unions, as for the European Union, price stability refers to the annual objective of keeping the Harmonized Index of Consumer Prices (HICP) within the Eurozone below 2\%. The existence of an anchor, as a quantitative target, is believed to be a powerful instrument for anchoring inflation expectations and facilitating the conduct of monetary policy strategies by central banks \citep{Castelnuovo_2003}. 

The HICP is an aggregation of prices collected within the Eurozone by the Eurostat. It serves two main purposes \citep{Eurostat_2024HICP}: quantifying the definition of price stability in the ECB monetary policy strategy and assessing price convergence for country candidates aiming at joining the monetary union.

Price collected at the Eurozone members are indeed heterogeneous and their patterns country-specific \citep{Eurostat_2024HICP}. Attempts to harmonizing price data were implemented at the methodology-level during mid-1970s and considerable effort were put later on developing the theory and practice of consumer price indices (CPIs).

Forecasters explored many econometric and machine learning methods to enhance the HICP predictability and successful results were obtained with the use of Deep Neural Networks (DNNs) \citep{Vancsura_2025}, confirming the difficulty encountered when using classic assumptions.


Country-level indices could be seen as a system of collected indices rendering an overview of price fluctuations within the Eurozone. This collection is indeed a system, where each country reports its index in a given period, yielding a matrix of data entries with a double dimension: country and time.

The study of systems has been first investigated in nuclear physics \citep{Wigner_1955}, where statistical properties of these system are essential to study their behavior. For this aim, many tools were derived from statistical mechanics, physics and probability to establish a dedicated brach of analysis: Random Matrix Theory (RMT) \citep{Potters_Bouchaud_2020}.

Generally, RMT deals with the behavior of eigenvalues, whose spectrum carries information about embedded patterns in large collections of data, or systems \citep{Ledoit_2011}. Eigenvalues are indeed random variables whose distribution depends solely on data properties. Classic applications considered the spectrum of eigenvalues and modeled it using Mar\v{c}enko-Pastur distribution \citep{Marchenko_1967} which plots a bell-like curve over eigenvalues used to capture the signal-noise duality in the original dataset. 

Further extensions of RMT were used to assess the goodness-of-fit of large systems, particularly DNNs \citep{Martin_2021, Martin_Mahoney_2021, Martin_2025}, by deriving eigenvalue-based quality metrics on data quality and system accuracy. Such tools have been applicable on weight matrices of DNNs, as well as at the data-level, on the basis of an information exponent, $\alpha$, that is empirically associated with optimal or ideal learning.

This works offers a genuine, novel perspective on inflation analysis by considering Eurostat inflation data as a system, instead of focusing on country-based price index. The latter is generally the weighted product of many sub-indices, hiding heterogeneous information that could be useful in interpreting specific developments and identifying local variations. RMT gives deeper insight on the asymptotic behavior of eigenvalues and offers extended analysis range, exceeding classic tools for multivariate time series whose scope mainly targets clustering and classification tasks \citep{Lopez_Oriona_2023}.

Applied on the monthly Eurostat inflation variations of 27 country members from January 2021 to July 2026, this approach confirmed the effect of small variations in preventing a clear separation of the signal from the noise. The Power-law index estimated on the Mar\v{c}enko-Pastur distribution, known as the \textit{information correlation}, shows a slight overfitting behavior of the system. This validates the persistent nature of small variations within many Eurozone countries, acting as a multivariate attention mechanism dominating the system. Despite this persistent behavior, the system demonstrates slightly overfit nuances close to perfect fit models, and reinforces the hypothesis of a self-organization state with improved robustness against errors.

Persistent small variations in the system further indicate the presence of modular structures in the correlation matrix. This evidence may hint hidden variability and inner clustered properties in the data. The spacing eigenvalues' distribution \citep{Menzel_2016} succeeded in learning a cutoff, under which correlations between country indices are assumed to be noise-driven. This denoising process identified two outliers and three distinct groups of countries with shared similarities in recent inflation developments. These findings suggest the use of regional-based indices to enhance the predictability of HICP, by taking into account a dynamic cluster formation of indices.

The remainder of the paper explains price stability from  a monetary perspective in Section \ref{section2}, then provides a theoretical framework to understand RMT tools (Section \ref{section3}), before setting the application on Eurostat data in Section \ref{section4} and a following discussion of results and suggestions (Section \ref{section5}).

\section{Price Stability}\label{section2}

The concept of price stability has a long history in modern macroeconomics and is still used as an anchor to guide strategies implemented by central banks and monetary committees around the world.

When conducting its monetary policy, a central bank has little control over short-term shocks affecting prices. It can devise a medium-term policy orientation to avoid excessive volatility in short-term interest rates or the real economy \citep{Smets_2003_2}. By doing this, the central bank shows a gradualist response to shocks that threaten price stability, which offsets such excessive volatility, while maintaining the price stability condition over the medium term 

\cite{Goodfriend_2001} preferred the term \textit{neutral policy} over price stability because it keeps output at its potential, with a constant markup of price over marginal cost. They recommended price stability as the core monetary strategy to ensure, among other conditions, the reduction of the output gap.

The Treaty on European Union gives the ECB the primary mandate of maintaining price stability over the medium run. Formulating the quantitative objective of price stability was left to the ECB's Governing Council. The latter defined price stability, during at a Governing Council meeting held in October 1998, as `` the year-on-year increase in the Harmonised Index of Consumer Prices for the euro area of below 2\%''.

Particularly in the Eurozone, price stability was tested using two main criteria: the stability of short- to medium-term inflation expectations and the absence of long-term price-level uncertainty \citep{Bordes_2007}.

Even if in many countries price stability represents a primary goal for monetary
policy, actual practices vary substantially across countries. They range from no explicit quantitative definition to explicit quantitative definitions and inflation targets (point targets or ranges for admissible inflation outcomes) \citep{Castelnuovo_2003}.

As a benchmark, the HCIP is supported by a set of legally binding standards related to the index quality, in addition to the coordination efforts of the Eurostat to harmonize compilation practices \citep{Eurostat_2026ecoicop2} across national statistical institutes\footnote{\url{https://www.ecb.europa.eu/stats/macroeconomic_and_sectoral/hicp/more/html/index.en.html}}. A Survey of Professional Forecasters was launched by the ECB in 1999 to help predicting main macroeconomic aggregates throughout the Euro area \citep{Kenny_2007} and have an expert assessment on future inflation dynamics.  

The HICP serves as the basis for many subsequent indices computed to help ECB monitoring inflation targeting, which differ in terms of methodology and data used \citep{Bodnar_2026}. For instance, \cite{Froehling_2022} computed a domestic inflation index for monetary policy transmission mechanism, based on import intensities of HICP and further information from national accounts and input-output tables. Further attempts to predict HICP used a multivariate approach as well as DNNs to enhance the predictability \citep{Vancsura_2025}, and confirmed issues due to the data heterogeneity.  

\section{Random Matrix Theory} \label{section3}

Data analysis was first a context-based discipline, meaning the tools used to extract insights were relevant to the field of application. One can cite statistics as an earlier approach to gain information from data, which led to the emergence of cross-disciplinary fields as for econometrics. 

The growing amount of data encouraged researchers to adopt advanced tools to maximize information extraction. While many practitioners still consider the idea of a \textit{model}, as a formalism of a solution addressing a given problem, physicists consider the collection of data as a system. The latter could also be seen as a blend of all parameters pertaining to a model, or as a collection of all data entries related to a given task. 

The behavior of a system, as for convergence and inner dependence, required advanced elements borrowed from statistical mechanics, physics and geometry. This gave birth to RMT as an inter-disciplinary field of research applying tools on different fields \citep{Potters_Bouchaud_2020}, ranging from nuclear physics to social sciences.

The statistical properties of a given system are retrieved by the analysis of its eigenvalues, as latent features bearing embedded information about data. This permits to study correlation dynamics and asymptotic properties for future directions as well as the duality signal-noise. 


While most datasets have a rectangular, tabular structure; the correlation matrix is used to extract eigenvalues and analyze their variability spectrum via the Mar\v{c}enko-Pastur distribution \citep{Marchenko_1967}. The latter describes, at the origin, the statistical properties of sample covariance matrices stemming from the Gaussian Orthogonal Ensemble \citep{Wigner_1955}. 

Initially, a T$\times$N matrix $\textbf{W}$ is assumed to have its elements $w_{ij}$ drawn from a normal distribution $\mathcal{N}(0,\sigma^{2})$. The Wishart transform, or the covariance matrix, is computed and have negligible non-diagonal elements. In practice, we use the sample correlation matrix on normalized inputs $\tilde{W}$, given by $\bm{X}=\frac{1}{N}\tilde{W}^\mathsf{T}\tilde{W}$ as the starting point for extracting eigenvalues $\lambda_{i}$. The eigenvalues spectrum on the original $W$ matrix has a probability density of Mar\v{c}enko-Pastur (MP): 

\[
f(\lambda) =
\begin{cases}
	\frac{N}{T}\frac{\sqrt{(\lambda_{+}-\lambda)(\lambda-\lambda_{-})}}{2\pi\sigma^{2}}    & \text{if $\lambda \in [\lambda_{-},\lambda_{+}]$ }, \\
	0 & \text{if $\lambda \notin [\lambda_{-},\lambda_{+}]$ }.
\end{cases}
\] 

where $\lambda_{-} = \sigma^{2}(1-\sqrt{\frac{T}{N}})^{2}$ and $\lambda_{+} = \sigma^{2}(1+\sqrt{\frac{T}{N}})^{2}$
 

The MP distribution considers the spectrum of eigenvalues bounded between $\lambda_{-}$ and $\lambda_{+}$ as representing the noise randomness, while eigenvalues falling outside the interval [$\lambda_{-}$,$\lambda_{+}$] are proxies of the signal. The MP paradigm is indeed a special case of the Wigner semicircle law \citep{Wigner_1955}, which stands for the asymptotic distribution of eigenvalues.

Because of embedded correlations in many systems, the $\bm{X}$ matrix is not strictly diagonal and the resulting MP distribution is somewhat deformed from its theoretical formula, due to data properties or learning schemes \citep{Martin_2025}. This deformation, known as heavy-tail, is an additional information on the difficulty to separate the signal from the noise in complex datasets. 

\cite{Martin_2021} devised an empirical evidence, called \textit{Heavy-Tailed Self-Regularization}, consisting on the use of a Power-law fit on the MP distribution to compute an \textit{information correlation} index $\alpha$, as an empirical goodness-of-fit measure of DNNs without accessing training and test data. Later, further constraints on eigenvalues distribution were brought in the Semi-Empirical Theory of Learning (SETOL) \citep{Martin_2025}. This extended the self-regularization theory \citep{Malevergne_2004}, which assumes the generic existence of a self-organized macroscopic state in any large multivariate system \citep{Sornette_2009}.

It was found \citep{Martin_2025} that nearly perfect-fit systems are related to $\alpha \sim 2$, while values in the range of [2,6] are synonyms of underfitting issues. Overfitting occurs with values below 2, which indicate a heavy-tailed distribution of eigenvalues with infinite sample variance due to memory patterns.

Researchers are usually interested in cleaning big correlation matrices and clustering their elements on specific subgroups. Depending on data complexity, these tasks could be performed if the noise has Gaussian properties and its matrix adds up to a deterministic matrix to reconstruct the original data \citep{Ding_2020}. Such schemes rely on Truncated Singular Value Decomposition (TSVD), which are not optimal when the noise exhibits departures from the Gaussian orthogonality. 

Inner clustered patterns, known as modular structures, emerge in complex datasets after eliminating noisy correlations, usually using a threshold to just keep relevant links. For this aim, the Nearest-Neighbor Spacing Distribution (NNSD) is used to study the probabilistic properties of ordered eigenvalues \citep{Menzel_2016}. 

For a spacing distribution $s_{i}=|\bar{\lambda}_{i}-\bar{\lambda}_{i-1}|$ of ordered eigenvalues $\bar{\lambda}_{i}$, the probability distribution related to pure random matrices is the Gaussian Orthogonal Ensemble $P_{GOE}(s) = \frac{\pi}{2}.s.exp(-\frac{\pi}{4}s^{2})$. The eigenvalues show a repulsive behavior, concentrated around a target far from $s=0$.

Modular structures exist when the entries of the original matrix are not random \citep{Menzel_2016}. This indicates a block-decomposition scheme following an exponential distribution $P_{Exp}(s) = exp(-s)$. The absence of repulsion in the exponential distribution helps finding clusters or subgroups in large networks and filter out noise-driven fluctuations. 

Genuine clusters could be retrieved via a candidate signal-noise separating threshold on the correlation matrix. The log-likelihood distances of the NNSD to both limiting distributions are iteratively calculated. The threshold is the point where a regime change, or a clear departure, occurs between both distances previously computed \citep{Menzel_2016}.

RMT tools often rely on specific metrics built around the eigenvalues spectrum. Considering ordered eigenvalues from highest to lowest values of a matrix $\textbf{X}$: $\bar{\lambda}_{1}>\bar{\lambda}_{2}>...>\bar{\lambda}_{n}>0$, we can compute the \textit{condition number} $\kappa_{E}(X)=\sqrt{\frac{\bar{\lambda}_{1}}{\bar{\lambda}_{n}}}$ to measure how sensitive a linear system is to small errors or noise \citep{Edelman_1988}. For $\kappa \sim 1$, the system is well-conditioned and low-sensitive, because its output remains stable when its input changes by a small amount. Alternatively, large values of $\kappa$ describes ill-conditioned systems with high sensitivity to noisy patterns.

\cite{Demmel_1988} proposed an alternative measure $\kappa_{D}(X)=\sqrt{\frac{\sum{\bar{\lambda}_{i}}}{\bar{\lambda}_{n}}}$ to test the difficulty of matrix inversion and to quantify the distance to the nearest ill-posed (singular) matrix. These measures may be used to test the ability of a system to amplify noisy-patterns and track their impact as new data entries are collected. Such iterative methods \citep{Weyl_1912} help quantify the degree of perturbation new datapoints may bring to exacerbate the system, or alternatively how robust are new data entries to the stability of the system.



\section{Application} \label{section4}

Monthly inflation rate variations of 27 Eurozone countries were gathered from the Eurostat website \footnote{\url{https://ec.europa.eu/eurostat/databrowser/product/page/prc_hicp_minr__custom_22494633}}, covering the period January 2001-July 2026. Data of these countries constitute a \textit{system} of prices, whose inner statistical properties will be investigated using RMT tools. Explicitly, the distribution of eigenvalues will be scrutinized to assess the signal-noise duality and the goodness-of-fit of the whole system. The interdependence of eigenvectors, through their spacing distribution, will be of a key importance to remove noise-related correlation which are not informative. This eases visualization and clustering of the countries, based on their truly informative links. 

The \textit{system} is a data matrix of 27 countries and 307 observations providing the per-country monthly evolution of inflation prices, as measured by the Consumption Price indices. No pre-processing steps or transformations on the system were conducted.

\subsection{SETOL}

The \textit{WeightWatcher} tool\footnote{\url{https://github.com/CalculatedContent/WeightWatcher}} is used to apply the SETOL methodology \citep{Martin_2025}. The eigenvalues spectrum is extracted and the empirical spectral density (ESD) is plotted and fitted using a log-log regression to get the information correlation $\alpha$.

\begin{figure}
	\centering
	\hspace*{-0.20cm}\includegraphics[scale=0.4]{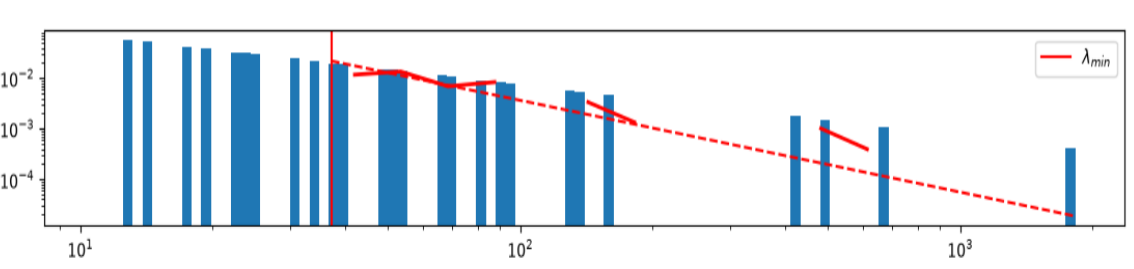}
	\caption{Log-Log Empirical Spectral Density and the Power-law fit for Eurozone prices, yielding $\alpha=1.82$ on the basis of $\lambda_{-}=37.21$} \label{fig1}
\end{figure}

Figure \ref{fig1} shows the Power-law fit on the basis of estimated $\lambda_{-}=37.21$ and $\sigma=0.2$, rendering an information correlation $\alpha=1.82$ with a Kolmogorov-Smirnov distance $D_{KS}=0.10$ computed vis-à-vis the theoretical Power-law. The value of $\alpha$ falls in the range of slighlty overfitted models, meaning many entries in the system are noise-dominated, mostly related to values below $\lambda_{-}$. The dispersion of the signal-related eigenvalues hints at potential inter-country correlations that hide inner clusters or modular structures. Knowing the limited size of the dataset, the aspect ratio $Q=\frac{T}{N}$ of the original matrix $W$ plays a significant role in the finite-size effect and the value of $\alpha$ \citep{Martin_2025}.

\begin{figure}[h]
 	\centering
 	\hspace*{-0.20cm}\includegraphics[scale=0.4]{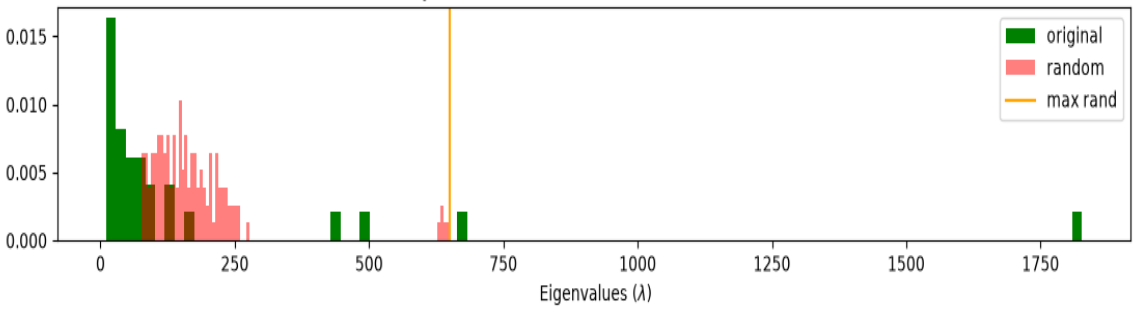}
 	\caption{Difference between the Log-Log Empirical Spectral Density of its theoretical (randomly generated) values (max rand ($\lambda_{+}^{rand}$) is the theoretical largest eigenvalue).} \label{fig2}
 \end{figure}

The price system cannot be assumed to be pure random. Figure \ref{fig2} displays a clear distinction between the empirical ESD from its random theoretical values. The system exhibits small departures in its eigenvalues distribution, called spikes, responsible of driving the value of $\alpha$ far from the ideal learning threshold ($\alpha \sim 2$). The spikes, large eigenvalues in Figure \ref{fig2} exceeding the $\lambda_{+}^{rand}$, reflect inflation fluctuations that depart from the sample. Such rare events cannot be assumed to be outliers, but are part of the system in what \cite{Sornette_2009} called \textit{Dragon-kings}. These \textit{meaningful outliers} distort the PL-fit of $\alpha$ but indicate the system is self-organized \citep{Sornette_2009}, having inner mechanisms exhibiting stability after episodes of high inflation.

The distance between the ESD and the spikes is considered as a phase-transition, or a regime transition that bifurcates the system, due to the economic status and internal/external factors exacerbating inflation pressures in some countries. These spikes are linked to persistent significant fluctuations recorded in some countries in the system $W$, rather than spuriously large elements in the correlation matrix $X$, a mechanism known as \textit{correlation trap} \citep{Martin_2025}.

The information correlation index $\alpha$, could be assumed to be a self-organization index \citep{Sornette_2009} and is indeed a measure of self-similarity, similar to the fractal dimension \citep{Mandelbrot_1975, Mandelbrot_1982} indicating the roughness of the ESD. \cite{Martin_2025} considers it as a proxy of implicit regularization, in constrast to explicit regularization used in most machine learning and DNNs.



 \begin{figure}[h]
	\centering
	\hspace*{-0.20cm}\includegraphics[scale=0.4]{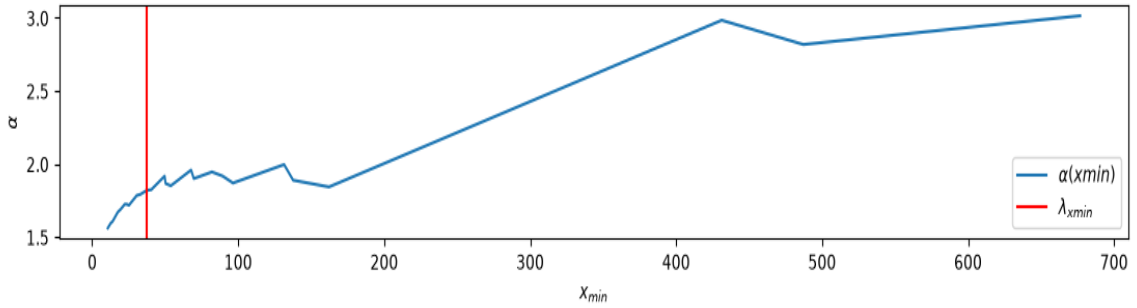}
	\caption{Variations of $\alpha$ values following different thresholds of $\lambda_{-}$. The selected $\lambda_{-}=37.21$ achieves the lowest Kolmogorov-Smirnov Distance ($D_{KS}$) when computing the Power-law fit.} \label{fig3}
\end{figure}

$\alpha$ is indeed a random variable, whose values depend on the determination of $\lambda_{-}$. Figure \ref{fig3} shows the spectrum of possible $\alpha$ values when different $\lambda_{-}$ are selected. The optimal $\lambda_{-}$ is selected on the basis of the lowest $D_{KS}$ between the log-log fit and the theoretical distribution. For values of $\lambda_{-}$ in the range of [100,160] the system could be assumed to have an efficient separation of the signal and noise, in other terms a \textit{perfect fit}. However, values greater than 200 demonstrate underfitting properties, where the system is not able to efficiently learn from the noise and the signal. In other terms, sidelining more eigenvalues will shrink the available information, resulting in increased values of $\alpha$ toward the underfitting status.

In probabilistic terms, some country-specific elements of the original matrix $W$ may have been drawn from a Power-law distribution rather than the normal distribution \citep{Martin_2025}, which explains the heavy-tailed empirical MP distribution.

The value of $\alpha$ of the price system is slightly below the perfect fit condition, and hints at a potential violation of the Mar\v{c}enko-Pastur hypothesis. In overfitted systems ($\alpha<2$) one may consider the initial data having an infinite variance and linked to a Power-law distribution, rather than a normal distribution. The goal is to better represent non-linearities of the entries $W_{ij}$, which result in heavy-tailed distribution of the eigenvalues. \cite{Martin_2025} suggested a Pareto distribution with an index (shape parameter) $\mu$, following: 

\[
w_{i,j}(\mu) \sim \frac{C}{x^{\mu+1}} \hspace{0.2cm};\hspace{0.2cm} \rho(\lambda) \sim \lambda^{-(a\mu + b)}
\] 

where $a=\frac{1}{2}$ and $b=1$ for ideal conditions ($\alpha \sim 2$) and $C$ a given constant. For a retrieved $\alpha=1.82$, one can assume the monthly inflation rates in the European Union as realizations of a Pareto distribution of $\mu=1.64$, confirming the non-linearities and the need of strong learners (DNNs) to predict HICP \citep{Vancsura_2025}.

\begin{figure}[h]
	\centering
	\hspace*{-0.15cm}\includegraphics[width=12cm]{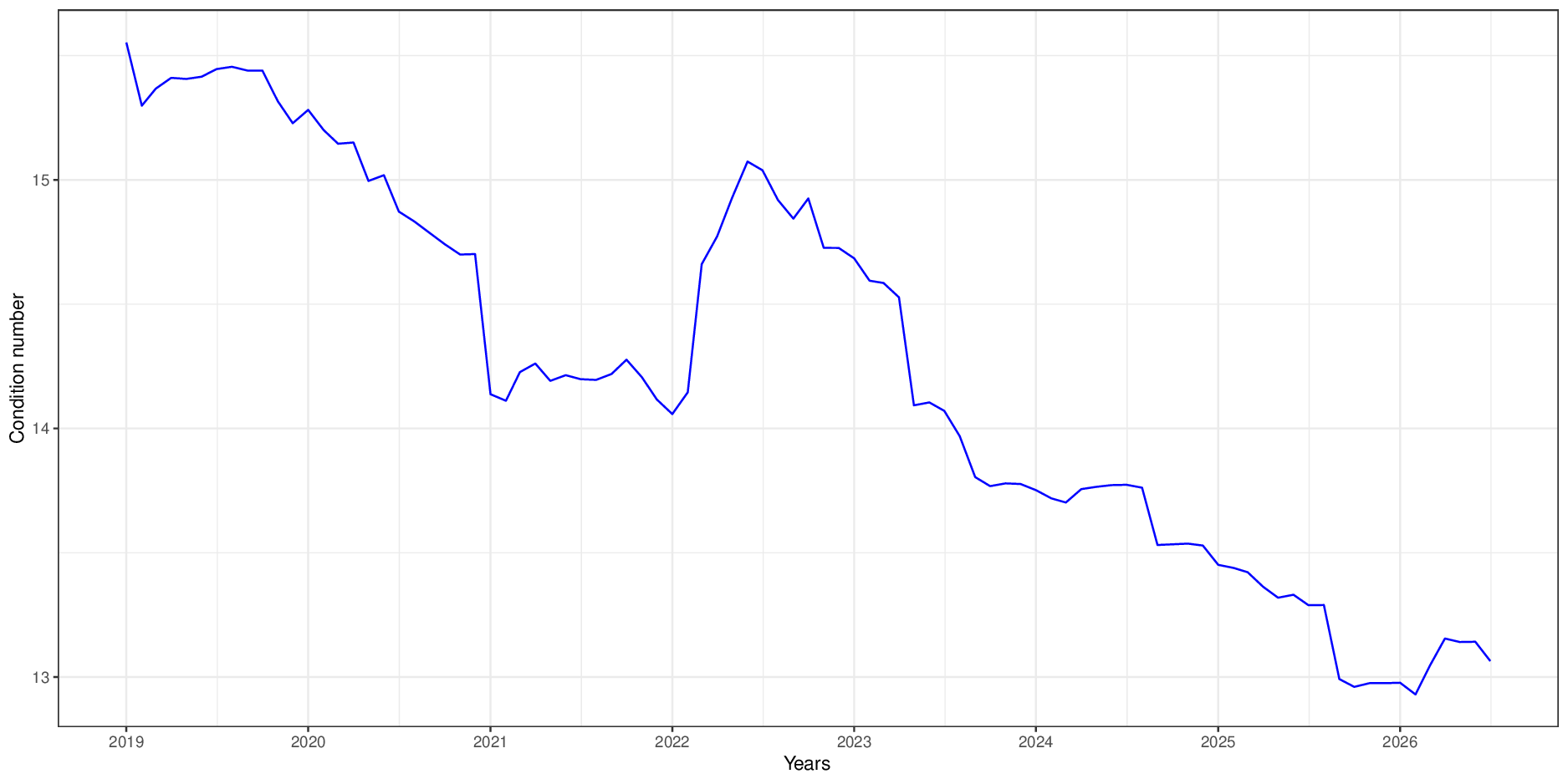}
	\caption{Plot of the iterative condition number during the period (2019-2026).} \label{fig4}
\end{figure}

Considering the period 2001-2018, the condition number $\kappa_{E}$ is iteratively computed on monthly new data entries in the system $\textbf{W}$ over the period 2019-2026. Figure \ref{fig4} shows a downward trend in the condition number values, except the year 2022, which indicates the system vulnerability to noisy patterns. This period witnessed generalized price jumps, where annual inflation rates in the European Union rose to double-digit numbers \citep{Ferreira_2025}. Despite the relative limited size of the sample, the system exhibits lower values of $\kappa_{E}$ meaning more robustness against errors and a certain systemic stability\footnote{The Demmel condition number series $\kappa_{D}$ is aligned to the standard $\kappa_{E}$ series with a correlation coefficient of 0.988.}, as the overall sample correlation (2019-2026) with the HICP was found to be at a level of -0.824.  

\subsection{Nearest-neighbor spacing distribution}

The country-based correlation matrix of the system, $\bold{X}$, is used to determine a threshold separating the signal from the noise. This denoising process aims to remove hidden, non-informative inter-country links pertaining to the noise. This allows to keep signals that help differentiate countries based on their inter-linkage and eventually, cluster their similarities in subgroups, or modular structures. 

\begin{figure}[bh]
	\centering
	\hspace*{-0.15cm}\includegraphics[scale=0.42]{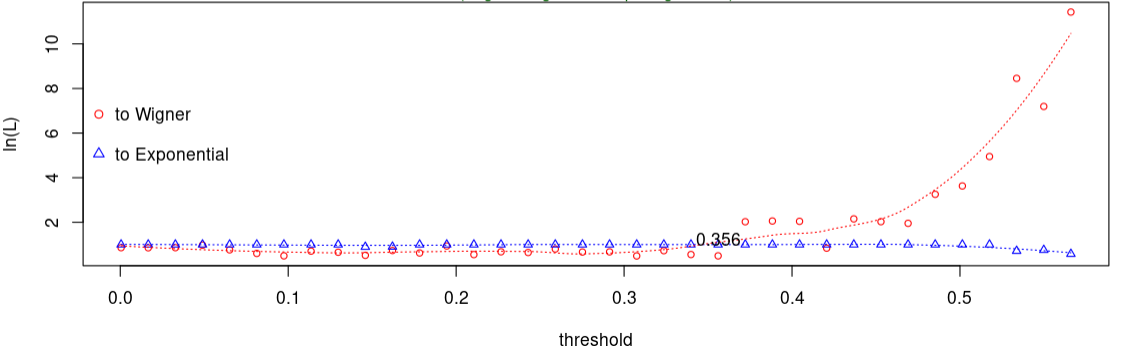}
	\caption{Plot of empirical negative log-likelihood distances to Wigner and Exponential distributions, following selected thresholds. The cutoff, or turning point, appears at the value of 0.356 .} \label{fig5}
\end{figure}

Figure \ref{fig5} shows the empirical NNSD negative log-Likelihood distances to both Wigner and exponential distributions. The point, at which both distances change their behavior, is assumed to be the threshold used to clean the correlation matrix \citep{Menzel_2016}. The cutoff of 0.356, a relatively higher positive correlation, appears to fulfill this condition and is next used to clean the original correlation matrix of the system to leave out links deemed to represent the noise.

Such transformations permit to ease clustering attempts to uncover modular structures, by considering significant correlation patterns in the system. Similar countries are clustered, in terms of communities, based on a deterministic analysis of short random walks \citep{Harel_2001} called \textit{Walktrap} \citep{Pons_2005} and leading eigenvector of the modularity matrix \citep{Newman_2006}. The latter is a spectral top-down approach, if compared to other bottom-up approaches as for Louvain algorithm \citep{Blondel_2008}.

\begin{figure}
	\centering
	\hspace*{-0.20cm}\includegraphics[scale=0.45]{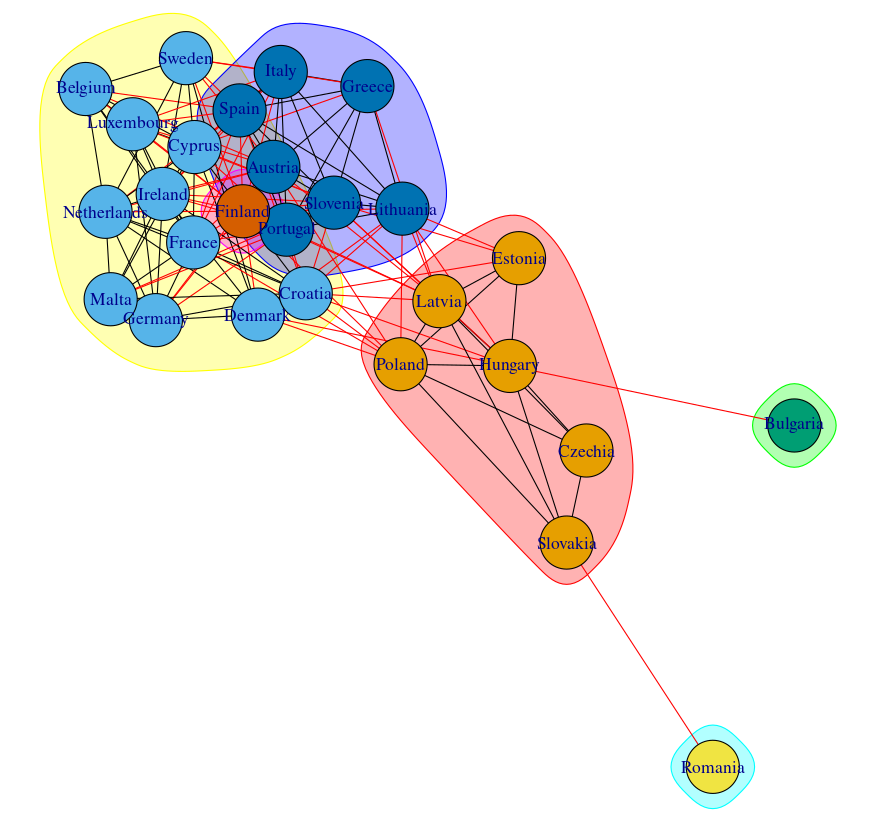}
	\caption{Communities reconstructed from the denoised country correlation matrix using leading eigenvector community detection \citep{Newman_2006}.} \label{fig6}
\end{figure}

The leading eigenvalue method (Figure \ref{fig6}) identified six distinct communities, based on their inner interactions throughout the study period. Romania and Bulgaria departs from other countries, having their own clusters, due to large inflation spikes in recent years. The third community clusters five east-European countries (Slovakia, Czechia, Hungary, Poland, Latvia and Estonia). Finland has its own community and lies between two other dense communities with geographical similarities.

The clustering offers interesting insights on regional inflation variability, responsible of fueling noisy patterns in the system. The integration of Romania and Bulgaria might have created spikes in the eigenvalues' distribution and moved away $\alpha$ from a perfect fit perspective. Based on the uncovered clusters, hidden patterns in the system cannot be imputed to two countries, but to other clusters of countries whose intermediary position might explain the difficulty of signal/noise separation and the higher cutoff value of the NNSD. 

Another factor that might have contributed to the clustering is the frequent update of the HICP basket for the sake of data quality. This hidden feature cannot be investigated at the country-level, but requires in-depth price collection and potential systemic analysis of the HICP divisions \citep{Eurostat_2026ecoicop2}.

\section{Discussion} \label{section5}

The use of RMT tools offered new perspectives in studying inflation dynamics in the European Union. Considering the whole system of monthly inflation rates of 27 countries, RMT identified latent non-stochastic patterns, confirming difficulties of predicting the HICP path over the medium run. The relatively small size of the sample does not permit to have robust generalizations, but showed strong evidences of small, but persistent cluster-driven deviations. Despite this hysteresis-effect, the system remains self-organized and assumes the spikes in the eigenvalues distribution as part of the sample, not as outliers.


Separating signal from the noise seems to be difficult in this particular dataset and needed a high positive threshold to filter out noise-related links, highlighting the non-stationary correlation dynamics in the system. The latter render modular structures facilitating a clustering of countries. The application of random walks community detection and further RMT clustering schemes confirmed the existence of regional communities with noticeable distinctions, based on recorded inflation spikes during last years.

The system is far from being random and the determination of stochasticity origins is quiet difficult, as it needs high-level data to identify more granular patterns. Several factors could have altered the system dynamics as for country-weight updates, cyclical inflationary trends and price methodologies performed to harmonize the collected data. 

Findings confirm the difficulty to predict HICP, based on existing price collection. The application of machine learning algorithms for predictive modeling will definitely need regularization schemes and transformations, that handle persistent small fluctuations. This challenging task could be easily performed with the help of synthetic data, as for \textit{Tabular Prior-Data Fitted Networks} (TabPFN) \citep{Hollmann_2022}, at the expense of explainability and interpretability needed by economists.

Clustering the system with the leading eigenvalue method yielded six distinct groups, highlighting the need to construct regional sub-indices to better apprehend the HICP path over the medium- and long-term. Aside from classic network clustering whose applications were unable to uncover clusters, the non-linear nature of the data entries required advanced RMT tools to get granular insights. 

Overall, the system has a self-organization feature similar to a mean-reverting behavior after the occurrence of large deviations from the sample distribution. The robustness against errors, as measured by the condition number, is another aspect of the system's stability and resilience despite the limited number of observations.

\section{Conclusion}

This paper offered a different way to analyze inflation at the European Union, by assuming inflation indices as components of a system rather than a multivariate dataset. This required advanced analytical tool to handle the cross-country price dynamics and inner correlations, seen as arbitrary structures with asymptotic properties. The system was found to be have persistent, small fluctuations that could be identified as noise-driven patterns similar to slight overfitted DNNs. This alters the separation of the signal from the noise, hence requiring solid inference techniques, especially for predictive exercises. The eigenvalues spectrum, as an informative feature of the system, exhibits heavy-tailed properties that distinguishes it from a pure random state. Overall, the system is dominated by small, memory-based deviations whose variability is group-based, rather than being country-related. The statistical properties of eigenvalues permits to cluster the system into six distinct groups of countries, indicating the need to compute sub-indices to better apprehend medium-term projections of the HICP and alleviate its variability. Despite non-linearities in the spectrum of eigenvalues, the system demonstrates a self-organizing state featuring large spikes that cannot be assumed to be outliers but part of the sample distribution. Self-organization is also a robustness that prevents the system from amplifying errors and reinforcing noisy-patterns. Further details on the HICP divisions will help getting more granular insights on inflation origins and potential cross-country trends, as persistency sources.

\newpage

\bibliographystyle{unsrt}
\bibliography{references}


\end{document}